\documentclass[conference]{IEEEtran}
\IEEEoverridecommandlockouts
\usepackage{cite}
\usepackage{hyperref}
\usepackage{amsmath,amssymb,amsfonts,dsfont}
\usepackage[linesnumbered,ruled,vlined]{algorithm2e}
\usepackage{graphicx}
\usepackage{textcomp}
\usepackage{svg}
\usepackage{multirow}
\usepackage{xcolor}

\usepackage[caption=false, font=footnotesize]{subfig}

\def\BibTeX{{\rm B\kern-.05em{\sc i\kern-.025em b}\kern-.08em
    T\kern-.1667em\lower.7ex\hbox{E}\kern-.125emX}}

\begin{document}

\title{Subject-Adaptive Sparse Linear Models for Interpretable Personalized Health Prediction from Multimodal Lifelog Data}
\author{%
  Dohyun Bu\IEEEauthorrefmark{2}\textsuperscript{*}, 
  Jisoo Han\IEEEauthorrefmark{3}\textsuperscript{*}, 
  Soohwa Kwon\IEEEauthorrefmark{3}\textsuperscript{*}, 
  Yulim So\IEEEauthorrefmark{4}\textsuperscript{*}, 
  Jong-Seok Lee\IEEEauthorrefmark{2}%
  \\[1ex]
  
  \IEEEauthorblockA{%
    \IEEEauthorrefmark{2}\textit{Department of Industrial \& Systems Engineering}, \textit{KAIST}, Daejeon, South Korea\\
  \IEEEauthorrefmark{3}\textit{Department of Systems Management Engineering}, \textit{Sungkyunkwan University}, Suwon, South Korea\\
  \IEEEauthorrefmark{4}\textit{Department of Industrial Engineering}, \textit{Sungkyunkwan University}, Suwon, South Korea\\[0.5ex]
    Emails:\ \{dohyun.bu, jongseok.lee\}@kaist.ac.kr;\ 
    \{jshan1004, sooah0201, you0715\}@skku.edu
  }

  \thanks{* These authors contributed equally to this work and are listed in alphabetical order.}
}

\maketitle

\begin{abstract}
Improved prediction of personalized health outcomes---such as sleep quality and stress---from multimodal lifelog data could have meaningful clinical and practical implications.
However, state-of-the-art models, primarily deep neural networks and gradient-boosted ensembles, sacrifice interpretability and fail to adequately address the significant inter-individual variability inherent in lifelog data. 
To overcome these challenges, we propose the Subject-Adaptive Sparse Linear (SASL) framework, an interpretable modeling approach explicitly designed for personalized health prediction. 
SASL integrates ordinary least squares regression with subject-specific interactions, systematically distinguishing global from individual-level effects. 
We employ an iterative backward feature elimination method based on nested $F$-tests to construct a sparse and statistically robust model. 
Additionally, recognizing that health outcomes often represent discretized versions of continuous processes, we develop a regression-then-thresholding approach specifically designed to maximize macro-averaged F1 scores for ordinal targets. 
For intrinsically challenging predictions, SASL selectively incorporates outputs from compact LightGBM models through confidence-based gating, enhancing accuracy without compromising interpretability. 
Evaluations conducted on the CH-2025 dataset---which comprises roughly 450 daily observations from ten subjects---demonstrate that the hybrid SASL-LightGBM framework achieves predictive performance comparable to that of sophisticated black-box methods, but with significantly fewer parameters and substantially greater transparency, thus providing clear and actionable insights for clinicians and practitioners.
\end{abstract}


\section{Introduction}\label{sec:intro}

Sleep is fundamental to physical health, emotional stability, and overall quality of life~\cite{seriessleepnet2023lee,sleep2021clement,the2024seighali}. 
Chronic sleep deprivation has been linked to severe health consequences, including hypertension, diabetes, obesity, depression, and heightened stress, which further exacerbates sleep issues, forming a detrimental cycle~\cite{stress2020martire,profiles2021robillard}. 
Thus, accurately monitoring and predicting both daily sleep quality and fatigue could deliver substantial clinical and health benefits.

Recent advancements in wearable technology and smartphone-based sensing have enabled continuous, fine-grained lifelogging, transforming everyday activities into high-resolution digital phenotypes~\cite{lifelogging2014gurrin,the2015jain,lifelog2022ribeiro}. 
These multimodal streams---including accelerometry, geolocation, app usage logs, and physiological signals---have shown significant promise in modeling and predicting emotional states, physical activities, sleep quality, and stress~\cite{a2021chriskos}. 

However, two key challenges remain: \textbf{(1) lack of interpretability} and \textbf{(2) inter-individual variability}. For lack of interpretability, state-of-the-art methods predominantly rely on deep neural networks and gradient-boosted ensembles~\cite{multi2024lee,multi2024cho,contextual2024park,tram2024kim,pixleepflow2024na}. Although these models can achieve strong predictive accuracy, their complexity often renders predictions \emph{uninterpretable}, limiting practical utility in clinical settings where understanding \emph{why} a prediction was made is crucial. For inter-individual variability, sensor–outcome relationships vary markedly across individuals; as a result, population-level models may miss subject-specific patterns, reducing accuracy and reliability in personalized contexts.

Furthermore, many health outcomes derived from lifelog data, such as sleep quality and fatigue levels, represent discretized forms of continuous states, naturally placing them within the scope of ordinal regression~\cite{conditional2014hothorn,deep2022kook}.
Treating such outcomes as standard classification problems may ignore their underlying continuous structure, while conventional regression methods fail to align with discrete evaluation metrics such as macro-averaged F1 (macro-F1), thus motivating a hybrid modeling approach.

To address these challenges, we propose the \textbf{Subject-Adaptive Sparse Linear} (\textbf{SASL}) framework. 
SASL retains ordinary least squares regression (OLS) at its core, ensuring interpretability, while incorporating only minimal complexity to achieve competitive predictive performance. 
The proposed framework explicitly models inter-individual variability through subject-specific interactions, differentiating global effects from personalized signals. 
An iterative nested $F$-test procedure~\cite{benchmark2020bommert}, designed to be statistically rigorous, ensures a parsimonious feature set by systematically removing statistically insignificant predictors.

Specifically, SASL employs a regression-then-thresholding strategy, first predicting continuous latent scores, and then deciding thresholds that maximize macro-F1 score, preserving the ordinal characteristics of the target variables. 
Moreover, recognizing that certain outcomes (e.g., sleep efficiency, S2) pose inherently greater modeling challenges, SASL selectively integrates predictions from a compact LightGBM~\cite{lightgbm2017ke} model, invoked via confidence-based gates, thereby improving overall accuracy without significantly compromising transparency.

Our key contributions are as follows:

\begin{itemize}
\item We present an interpretable, statistically rigorous, subject-adaptive linear modeling approach that leverages sparse and carefully selected predictors to distinguish global from subject-specific effects.
\item We propose a regression-then-thresholding strategy for ordinal targets that is specifically aligned with macro-F1 evaluation.
\item We develop a selective hybrid scheme that activates compact black-box components only for intrinsically challenging predictions, thereby balancing transparency with state-of-the-art accuracy.
\end{itemize}

The remainder of this paper is organized as follows. Section~\ref{sec:re} reviews related work. Section~\ref{sec:me} presents the proposed SASL framework. Section~\ref{sec:eval} reports the empirical evaluation on the CH-2025 dataset~\cite{ETRI2025_Lifelog2024} and the interpretability analyses. Finally, Section~\ref{sec:con} concludes.

\section{Related Works}\label{sec:re}

\subsection{Sensor-based Health Monitoring}
Recent advances in mobile and wearable sensors have enabled continuous, unobtrusive monitoring of human behavior and physiology in natural settings. 
Multimodal lifelog data---comprising physical activity (e.g., accelerometer, step count), physiological signals (e.g., heart rate), environmental context (e.g., light, GPS), and mobile usage patterns---offer a rich foundation for modeling sleep, fatigue, and stress.

Prior studies have applied time-series models to capture temporal dependencies in lifelog data, addressing prediction tasks such as cognitive impairment, app churn, and stress detection~\cite{kim2024prediction, elbasani2021llad, kwon2021churn}. 
Transformer-based architectures have demonstrated strong performance in modeling long-range dependencies and integrating heterogeneous modalities~\cite{nguyen2024concept, park2025tram}. 
In addition, gradient-boosted ensembles have been employed to model health indicators such as blood glucose and sleep quality from contextual features~\cite{choi2023glucose, jung2024sleep}. 
CNN-based approaches have also been explored, converting sensor signals into structured or visual representations to enhance activity recognition and anomaly detection~\cite{kim2023abnormal, lee2022multimodal}.
However, many of these advances rely on black-box architectures, which can make interpretability challenging and may limit clinical trust, actionable insight, and subject-specific personalization.

\subsection{Statistical Modeling}
Statistical modeling has played a central role in health prediction tasks involving behavioral and physiological data, offering interpretable structures and well-defined inference procedures. 
Ordinary least squares (OLS) regression~\cite{rao1973ols}, which models linear relationships by minimizing squared errors, has been widely used in tasks such as fatigue monitoring and sleep quality estimation, where model transparency and simplicity are essential~\cite{benchmark2020bommert}.
To improve flexibility while retaining interpretability, recent studies have introduced extensions such as mixed-effects models and subject-specific interaction terms~\cite{conditional2014hothorn}. 
These methods allow statistical models to account for individual variation while maintaining a global structure.

In addition, various feature elimination strategies have been proposed to enhance statistical robustness in high-dimensional settings. 
Among them, backward elimination procedures based on nested $F$-tests have been used to construct sparse linear models with reliable explanatory power~\cite{graph2023chen}. 
For prediction problems involving ordinal targets, regression-then-thresholding frameworks have also gained attention, as they align naturally with metrics such as the classification accuracy~\cite{frank2001simple}.

Taken together, these strands suggest that integrating expressive black-box models with sparse, subject-adaptive linear components and ordinal-aware decision rules may yield accurate, transparent predictions from lifelog data.

\section{Method}\label{sec:me}
Our proposed SASL pipeline consists of five primary stages: (\ref{subsec:data}) data preprocessing and full model specification; (\ref{subsec:back}) backward feature elimination; (\ref{subsec:time}) time-aware evaluation and inference; (\ref{subsec:special}) special treatment for S2; and (\ref{subsec:conf}) confidence-based post-processing. Collectively, they balance accuracy and transparency by coupling sparse, subject-adaptive OLS with leakage-aware validation, ordinal-aware thresholds, and a narrowly applied, high-confidence LightGBM assist.

\subsection{Data Preprocessing and Full Model Specification}\label{subsec:data}
\vspace{-5pt}
We first summarize each daily observation using interpretable statistical features. 
These include one-hot encoded weekdays (Mon, Tue, Wed, Thu, Fri, Sat, Sun), a numeric day-of-week index (\texttt{dow}: Monday=0 to Sunday=6), the ratio of screen-on time per day, calories expended per day, smartphone charging duration per day, and the proportions of each smartphone-detected activity type (e.g., driving, walking, running, stationary). 
All continuous variables are standardized for stable linear regression fitting.

To model inter-individual variability, subject identifiers (\texttt{id1} to \texttt{id10}) are encoded using one-hot vectors. 
We introduce interaction terms between subject IDs and each statistical feature, explicitly capturing subject-specific deviations. 
For instance, a single observation initially encoded as \texttt{\{subject id, screen-on ratio\}} is augmented to include \texttt{\{id1, id2, screen-on ratio, id1 $\times$ screen-on ratio, id2 $\times$ screen-on ratio\}}, allowing the regression model to estimate personalized coefficients for each individual. 
The resulting full model is equivalent to fitting separate linear models for each subject.

\subsection{Backward Feature Elimination}\label{subsec:back}
\vspace{-5pt}
To prune the inflated design matrix produced by the subject–interaction terms, we adopt an \emph{iterative nested $F$–test} elimination scheme detailed in Algorithm~\ref{alg:backward_f}. Here, $F_{\!\mathrm{cdf}}(\cdot)$ is the cumulative distribution function of an $F$-distribution, $\mathrm{cols}(\mathbf X)$ returns the number of columns of matrix $\mathbf X$, and $\operatorname{randChoice}(\mathcal C, \cdot)$ randomly selects one index from $\mathcal C$.
The procedure begins with the full model and evaluates each candidate predictor by contrasting a reduced specification---obtained by omitting that single term---against the current model via the statistic
\begin{equation}
  F \;=\;
  \frac{\bigl(\mathrm{SSE}_{\text{reduced}}-\mathrm{SSE}_{\text{full}}\bigr) / 1}
       {\mathrm{SSE}_{\text{full}}\,/\,\bigl(N-p\bigr)},
  \label{eq:nested_f}
\end{equation}
where $\mathrm{SSE}$, $N$, and $p$ denote the residual sum of squares, the number of observations, and the number of features in the full design matrix. 
The feature whose associated $p$–value exceeds the significance threshold~$\alpha$ by the greatest margin is removed. 
The model is subsequently refitted, and the test is repeated. 
Ties---caused chiefly by multi-collinearity---are resolved using randomized priority, after which the resulting specification is vetted on validation data; the tie-breaking seed that maximizes ROC-AUC is retained. 
The loop stops once no term can be removed without violating the $\alpha$ criterion, yielding a statistically parsimonious predictor set $\mathbf X_{\text{reduced}}$ and an audit trail $\mathcal H$ of all deletion steps.

\begin{algorithm}[h]
\caption{Backward Elimination via Nested $F$-Tests}
\label{alg:backward_f}
\DontPrintSemicolon
\SetKwFunction{SSE}{SSE}
\SetKwFunction{Cols}{cols}
\KwIn{$\mathbf X_{\text{full}}\in\mathbb R^{N\times p},\; \mathbf y\in\mathbb R^{N},\; \alpha, \;s$}
\KwOut{$\mathbf X_{\text{reduced}}$}

\BlankLine
\textbf{Helper:} \textit{SSE}($\mathbf X$,$\mathbf y$) returns the SSE of the OLS fit on $\{\mathbf x_i, y_i\}_{i=1}^N$ where $\mathbf [\mathbf x_i{:}y_i] \in [\mathbf X{:}\mathbf y]$.\;

\BlankLine
$\mathbf X \gets \mathbf X_{\text{full}}$,\quad $\mathcal H \gets \emptyset$,\quad $t \gets 0$\;

\vspace{0.25em}

\While{\Cols{$\mathbf{X}$} $> 1$}{
  $\text{SSE}_{\text{full}} \gets$ \textit{SSE}\text{(}{$\mathbf{X},\mathbf{y}$}\text{)}\;
  $\text{df}_{\text{full}} \gets N - \text{\Cols}{\mathbf{X}}$\;

\vspace{0.25em}

  \ForEach{predictor index $j$ in $\mathbf X$}{
    $\mathbf{X}_{-j} \gets \mathbf{X}$ without column $j$\;

\vspace{0.25em}
    
    $\text{SSE}_{\text{reduced}}^{(j)} \gets$ \textit{SSE}\text{(}{$\mathbf X_{-j},\mathbf y$}\text{)}\;

\vspace{0.25em}
    
    $F_j \gets \dfrac{\bigl(\text{SSE}_{\text{reduced}}^{(j)}-\text{SSE}_{\text{full}}\bigr)/1}{\text{SSE}_{\text{full}}/\text{df}_{\text{full}}}$

\vspace{0.25em}
    
    $\text{pval}_j \gets 1 - F_{\mathrm{cdf}}(F_j;\,1,\,\text{df}_{\text{full}})$\;
  }

\vspace{0.5em}

  $\mathcal C \gets \{\,j : \text{pval}_j = \max_j\, \text{pval}_j\,\}$\;
  $j^\star \gets \operatorname{randChoice}(\mathcal C,\,\text{seed}=s)$

\vspace{0.25em}

  \eIf{$\max_j\, \text{pval}_j > \alpha$}{
    $t \gets t+1$\;
    $\mathbf X \gets \mathbf X_{-j^\star}$\;
  }{
    \textbf{break}\;
  }
}

\Return{\ensuremath{\mathbf{X}_{\text{reduced}} \gets \mathbf{X}}}
\end{algorithm}

\begin{figure}[h]
  \centering
  \includegraphics[width=\linewidth]{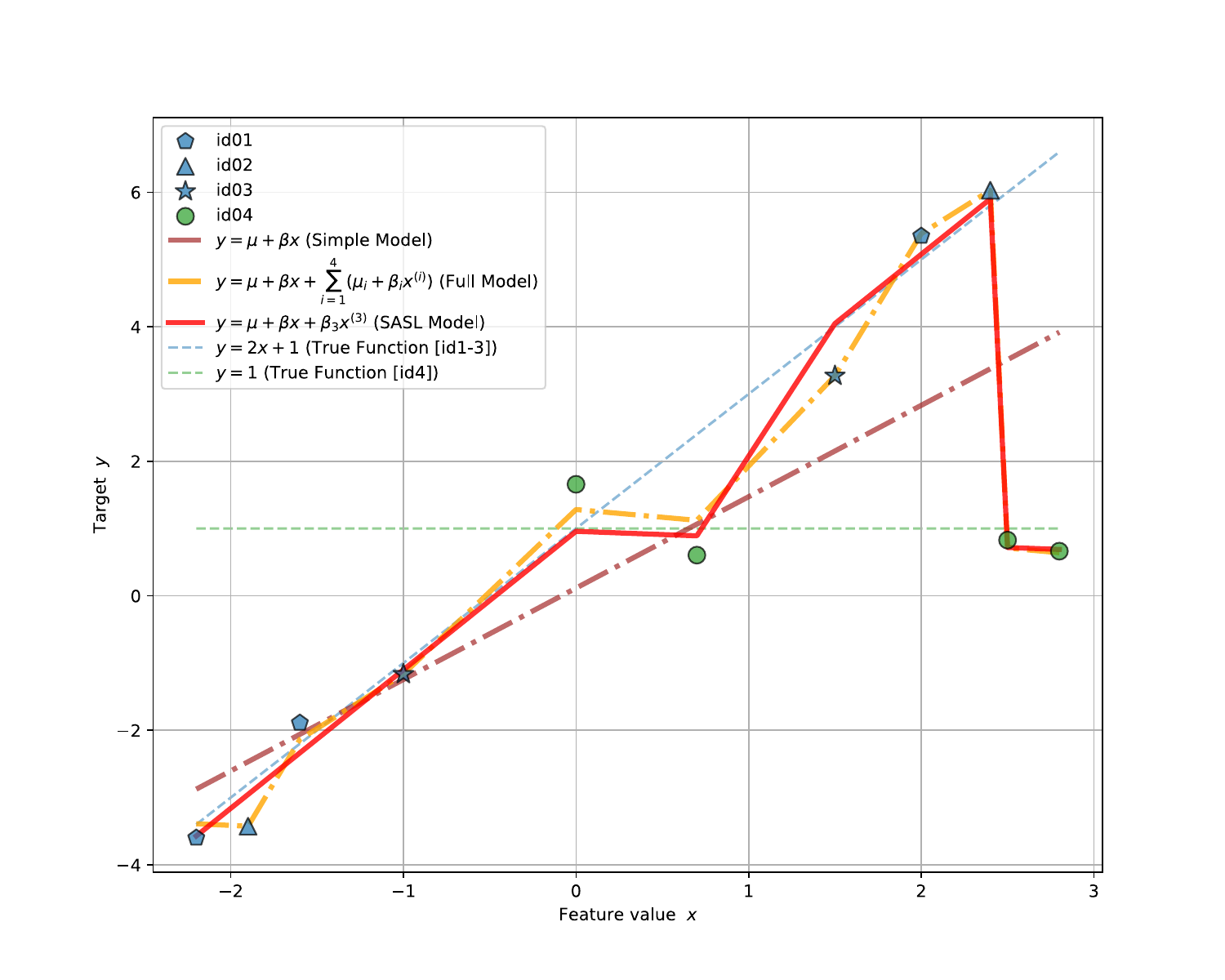}
  \caption{%
    Toy regression example after backward feature elimination. SASL (red solid) preserves a single subject-specific slope $\beta_{3}$ while discarding the remaining subject ID interactions, striking a compromise between the under-fitted global model $y=\mu+\beta x$ (brown dash-dot) and the over-parameterized full model (orange dashed dot). Here $\mu$ and $\beta$ denote the \emph{global} intercept and coefficient shared by all subjects, whereas $\mu_{i}$ and $\beta_{i}$ are the additional intercept and slope for subject~$i$. The term $x^{(i)}$ represents the interaction between the feature $x$ and the one-hot indicator for subject~$i$, meaning that it equals $x$ for that subject and $0$ for others.
  }
  \label{fig:sasl_toy}
\end{figure}

Figure \ref{fig:sasl_toy} visualizes how the backward $F$-test pares the full design matrix down to a sparse specification that keeps only statistically indispensable subject-specific terms. The simple global model, the unreduced full model, and the SASL fit are shown against the underlying ground-truth functions for the four synthetic subjects.

\vspace{-15pt}
\vspace{-5pt}

\subsection{Time-aware Evaluation and Inference}\label{subsec:time}
\vspace{-5pt}
As sleep- and fatigue-related outcomes exhibit strong temporal continuity, random shuffling would introduce temporal leakage into training, biasing the evaluation and overstating performance. 
To avoid this, we employed a chronological two-fold cross-validation scheme. 
Specifically, each subject’s timeline was split at its median timestamp, yielding two segments: an early segment and a late segment. 
In fold-1, we trained on early segments and validated on late segments across all subjects; fold-2 reversed these roles. 
Thus, each fold preserved temporal ordering and subject balance, enabling a realistic assessment of model performance.

During backward feature elimination, multiple predictors occasionally exhibited equally non-significant effects, causing ties. 
To resolve such ties systematically, we introduced a priority-tuning step. 
We considered a set of random seeds $\mathcal{S}$, each defining a unique priority order for tie-breaking among predictors. 
For each seed $s \in \mathcal{S}$, we first applied backward feature elimination to the entire training set, obtaining a reduced feature subset $\mathbf{X}_{\text{reduced}}^{(s)}$. 
Next, for each chronological fold $f\in\{1,2\}$, we fit an OLS model using the reduced subset, evaluated performance via ROC-AUC on the validation split of the corresponding fold, and computed the seed-specific average performance:
\begin{equation}
\text{Score}(s)=\frac{1}{2}\left(\text{ROC\_AUC}_{\text{fold-1}}^{(s)}+\text{ROC\_AUC}_{\text{fold-2}}^{(s)}\right).
\end{equation}
The final seed $s^\star$ was selected as
\begin{equation}
s^\star = \arg\max_{s\in\mathcal{S}}\; \text{Score}(s),
\end{equation}
producing a deterministic and performance-optimized reduced feature set for subsequent modeling.

With the feature subset fixed by $s^\star$, we then implemented a regression-then-thresholding strategy specifically aligned with the macro-F1 metric. 
For each candidate threshold $\tau$ (or threshold pair $(\tau_1, \tau_2)$ for ternary targets), we separately computed the macro-F1 score on each chronological fold. 
Rather than directly selecting the threshold that maximizes the average macro-F1, we searched for a stable plateau region where both folds maintained consistently high macro-F1 scores. 
Formally, for a binary target, the discretization was defined as:
\begin{equation}
\hat{y} = \mathds{1}\left[\hat{z}>\tau^\star\right],
\end{equation}
where $\hat{z}$ denotes the OLS prediction, and $\tau^\star$ denotes the chosen robust threshold. For the ternary target (S1), the discretization was defined as:
\begin{equation}
\hat{y} = \mathds{1}\left[\hat{z}>\tau_1^\star\right] + \mathds{1}\left[\hat{z}>\tau_2^\star\right],
\end{equation}
where thresholds $\tau_1^\star$, $\tau_2^\star$ were likewise selected based on plateau stability.

Finally, a single OLS model was fit on the entire training set, using the finalized reduced feature set to estimate the regression coefficients $\widehat{\beta}^{(s^\star)}$. 
Test instances were scored as:
\begin{equation}
\hat{z}=\left(\widehat{\beta}^{(s^\star)}\right)^\top \mathbf{X}_\text{test},
\end{equation}
which were then discretized using the selected robust thresholds. 
These final predictions were then refined via a confidence-based gating step, incorporating predictions from a complementary LightGBM classifier, yielding the competition submission.

\subsection{Special Treatment for \textbf{S2} (Sleep Efficiency)}\label{subsec:special}
\vspace{-5pt}
Unlike the other five targets, sleep efficiency (S2) captures nocturnal-only signals that are difficult to infer accurately from the coarse day-level summaries used in SASL. 
To address this inherent complexity, we developed a dedicated pipeline incorporating detailed temporal features and a robust modeling strategy specifically for S2. 
Minute-level sensor data were first re-indexed into consistent ``analysis days'' spanning from 16:00 to the following day's 16:00 (KST) to align with individual sleep cycles. 
After appropriate domain-specific imputations, we computed 172 daily statistical aggregates, including interaction terms (e.g., \textit{heart\_rate}$\times$\textit{steps}), individual-specific baseline deviations, and event flags indicating nocturnal disturbances (\textit{night\_bright}). 
Additionally, subjects were grouped into 10 behavioral archetypes using $k$-means clustering on key routine features, which were then encoded as categorical features.

To manage feature dimensionality, we applied Recursive Feature Elimination (RFE) with a LightGBM classifier, reducing the input set to the top 30 informative features. 
The final model was trained via stratified 5-fold cross-validation repeated over multiple seeds (20 total folds), incorporating early stopping based on validation ROC-AUC scores to mitigate overfitting. 
Optimal classification thresholds maximizing macro-F1 scores were identified separately within each fold; the final threshold was set to the mean of the fold-specific values.

\subsection{Confidence-based Post-processing}\label{subsec:conf}
\vspace{-5pt}
To complement the linear model's interpretability with nuanced non-linear relationships, we implemented a confidence-based post-processing step that leverages predictions from a LightGBM classifier. 
Confidence scores were computed as the maximum predicted class probability output by the LightGBM model.
When predictions from SASL and LightGBM models differed, we selectively adopted the LightGBM model's prediction only if it demonstrated exceptionally high confidence ($\geq$ 0.97 for all targets except S2, where $\geq$ 0.943 was used). 
This hybrid approach preserved overall stability and interpretability from the linear model while correcting select predictions where the LightGBM classifier demonstrated strong confidence, thereby modestly enhancing final predictive performance.

\section{Evaluation}\label{sec:eval}

Experiments were conducted on the public \textbf{CH-2025} benchmark dataset~\cite{ETRI2025_Lifelog2024}, which consists of approximately 450 daily records, each summarized into a 2,160-dimensional feature vector across ten subjects. 
Following the time-aware evaluation approach described earlier (Section~\ref{subsec:time}), we constructed two chronological folds for each subject by splitting their timeline at the median timestamp. 
One fold trained on early segments and validated on later segments, while the other reversed this order. 
All model eliminations---including feature elimination, priority tuning, and threshold optimizations---were performed strictly within these folds, ensuring no leakage of temporal information into training. 
The final performance results are based on the competition's held-out public test set, evaluated exactly once. 
Macro-F1 was computed on this public test split.

Table~\ref{tab:overall_results} summarizes the key results. 
The baseline SASL framework alone achieved a macro-F1 of 0.6258. 
Incorporating a dedicated LightGBM component for the most challenging target (S2, sleep efficiency) increased performance to 0.6311. 
A further improvement to 0.6387 was attained by selectively integrating high-confidence LightGBM model predictions for other targets as part of a confidence-based post-processing step. 
This final hybrid approach secured the fourth position among 1,032 participants on the competition's public leaderboard.

\begin{table}[h]
  \centering
  \caption{Macro-F1 Performance Comparison (Public Test Score)}
  \label{tab:overall_results}
  \begin{tabular}{l c}
    \hline
    Method & macro-F1 \\
    \hline
    SASL only                & 0.6258 \\
    SASL + LightGBM(S2)              & 0.6311 \\
    SASL + LightGBM(S2) + Post-processing       & \textbf{0.6387} \\
    \hline
  \end{tabular}
\end{table}

The optimized decision thresholds identified through the plateau search (Section~\ref{subsec:time}) exhibited strong temporal robustness. 
Specifically, we adopted thresholds of $\tau_{Q1}{=}0.424$, $\tau_{Q2}{=}0.578$, $\tau_{Q3}{=}0.603$, $\tau_{S3}{=}0.650$ for binary outcomes, and $\tau_{1}{=}0.900$, $\tau_{2}{=}1.125$ for the ternary outcome S1.

We closely examined sleep efficiency (S2), an inherently challenging target, by analyzing disagreements between the SASL and the LightGBM classifier. 
Out of 250 test-day predictions, the two models disagreed on 71 instances. The Z-score $Z$ is calculated to measure the deviation of a feature's mean within a specific group $\mu_{\text{group}}$ from the global mean $\mu_{\text{global}}$ of the entire dataset, standardized by global standard deviation $\sigma_\text{global}$. The formula is expressed as:

\begin{equation}
Z = \frac{\mu_{\text{group}} - \mu_{\text{global}}}{\sigma_{\text{global}}}
\end{equation}

\begin{figure}[h]
  \centering
  \subfloat[\textit{LightGBM}(0),\; \textit{SASL}(1)\label{fig:s2_z_low}]{%
    \includegraphics[width=0.48\linewidth]{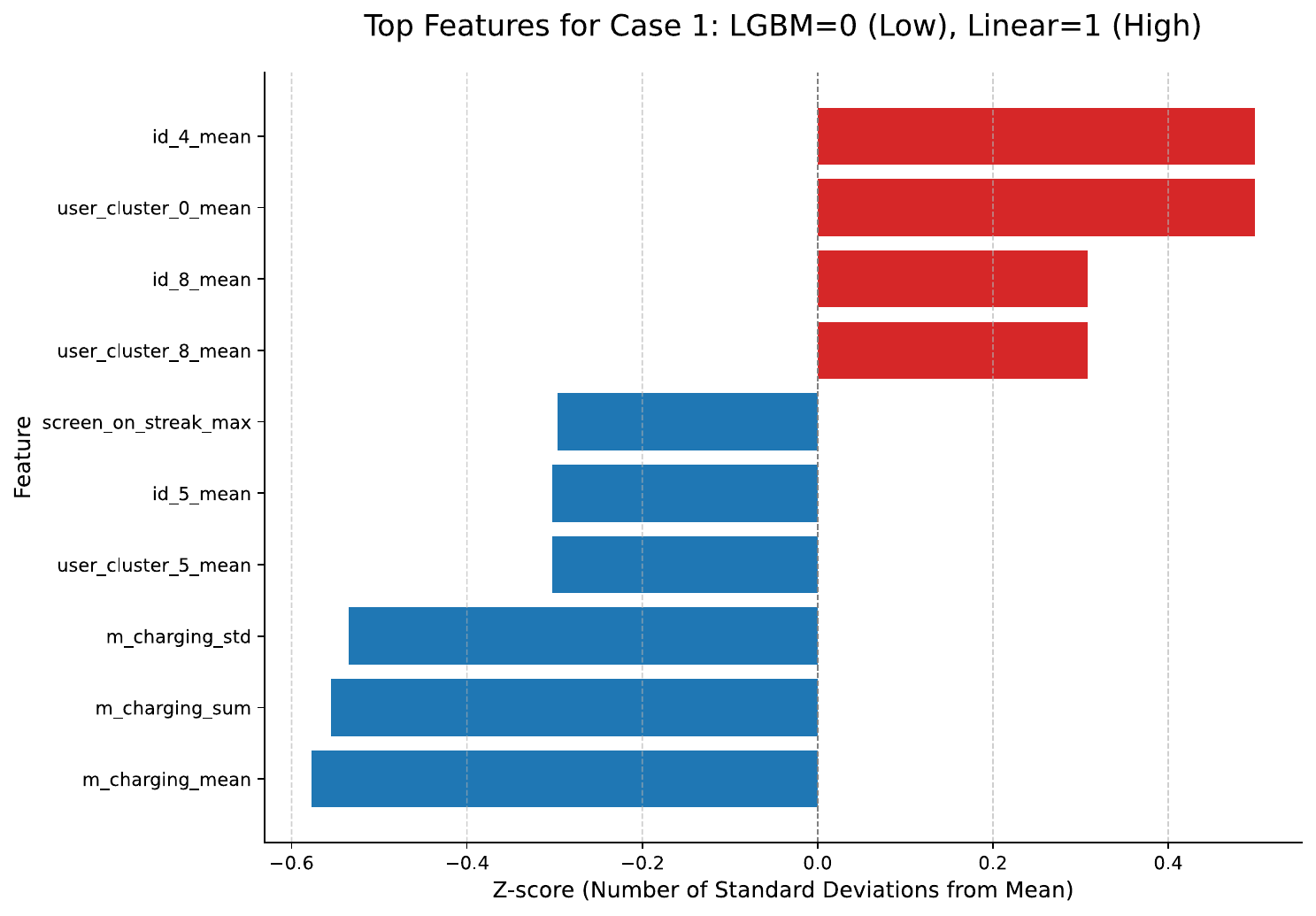}}
  \hfill
  \subfloat[\textit{LightGBM}(1),\; \textit{SASL}(0)\label{fig:s2_z_high}]{%
    \includegraphics[width=0.48\linewidth]{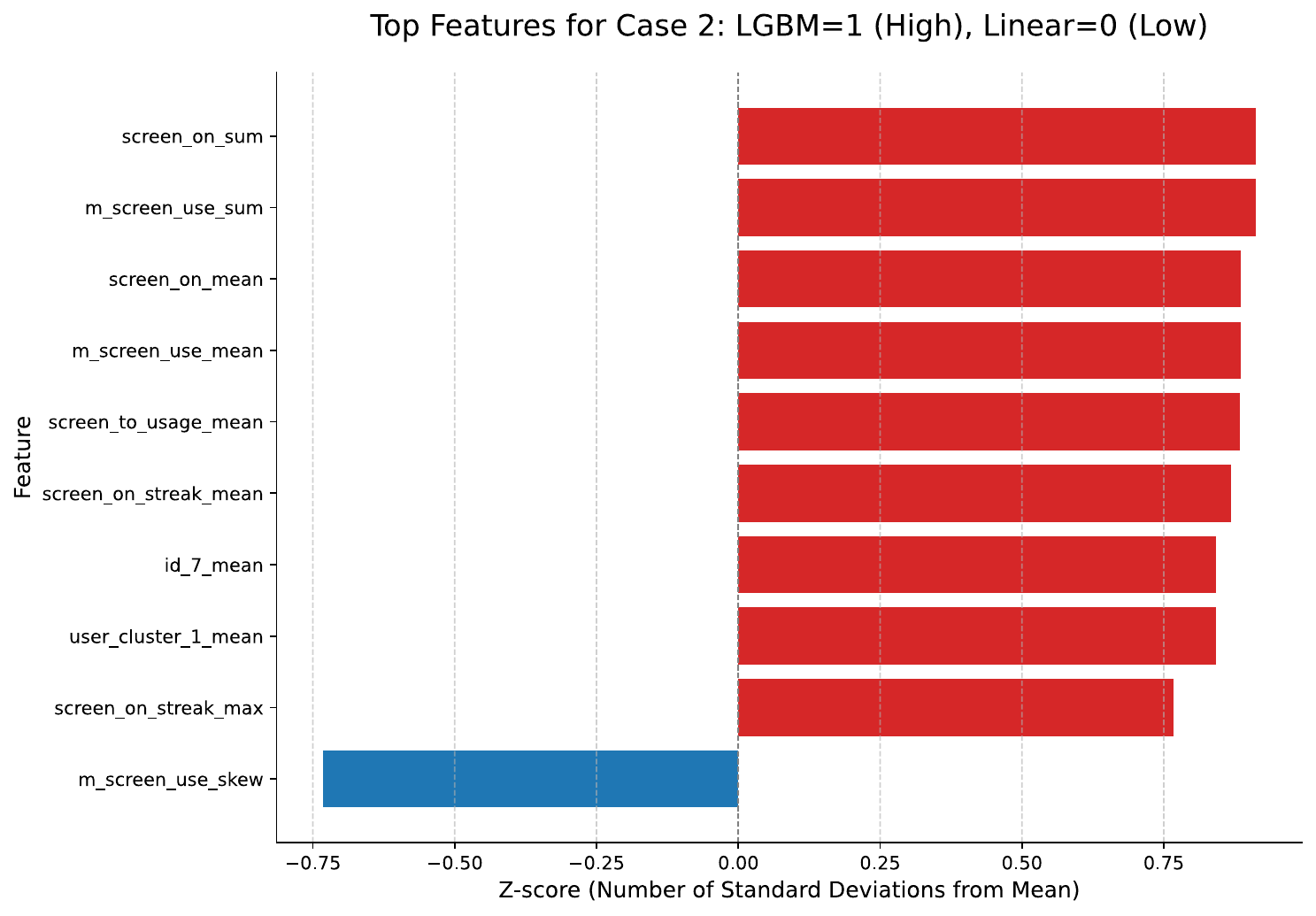}}
  \caption{Z-score profiles for the two disagreement groups.}
  \label{fig:z_score_profiles}
\end{figure}

By computing standardized Z-scores of feature averages within these conflicting cases, we identified clear behavioral patterns driving discrepancies. 
When the LightGBM classifier predicted low efficiency (0) but SASL predicted high (1) (Figure \ref{fig:s2_z_high}), the main signal was disrupted smartphone charging routines (e.g., low \texttt{m\_charging\_mean}; Z-score: $-0.58$). 
Conversely, when SASL predicted low efficiency and LightGBM high (Figure \ref{fig:s2_z_low}), discrepancies were dominated by excessive screen usage (e.g., high \texttt{screen\_on\_sum}; Z-score: $+0.91$). 
This implies that the non-linear LightGBM model successfully captured nuanced contexts---routine disruptions and compensatory behaviors---that the linear SASL overlooked or over-penalized.

\vspace{-10pt}
\vspace{-2pt}

\begin{figure}[h]
  \centering
  \subfloat[Q1\label{fig:q1_coef}]{%
    \includegraphics[height=3.3cm, width=8cm]{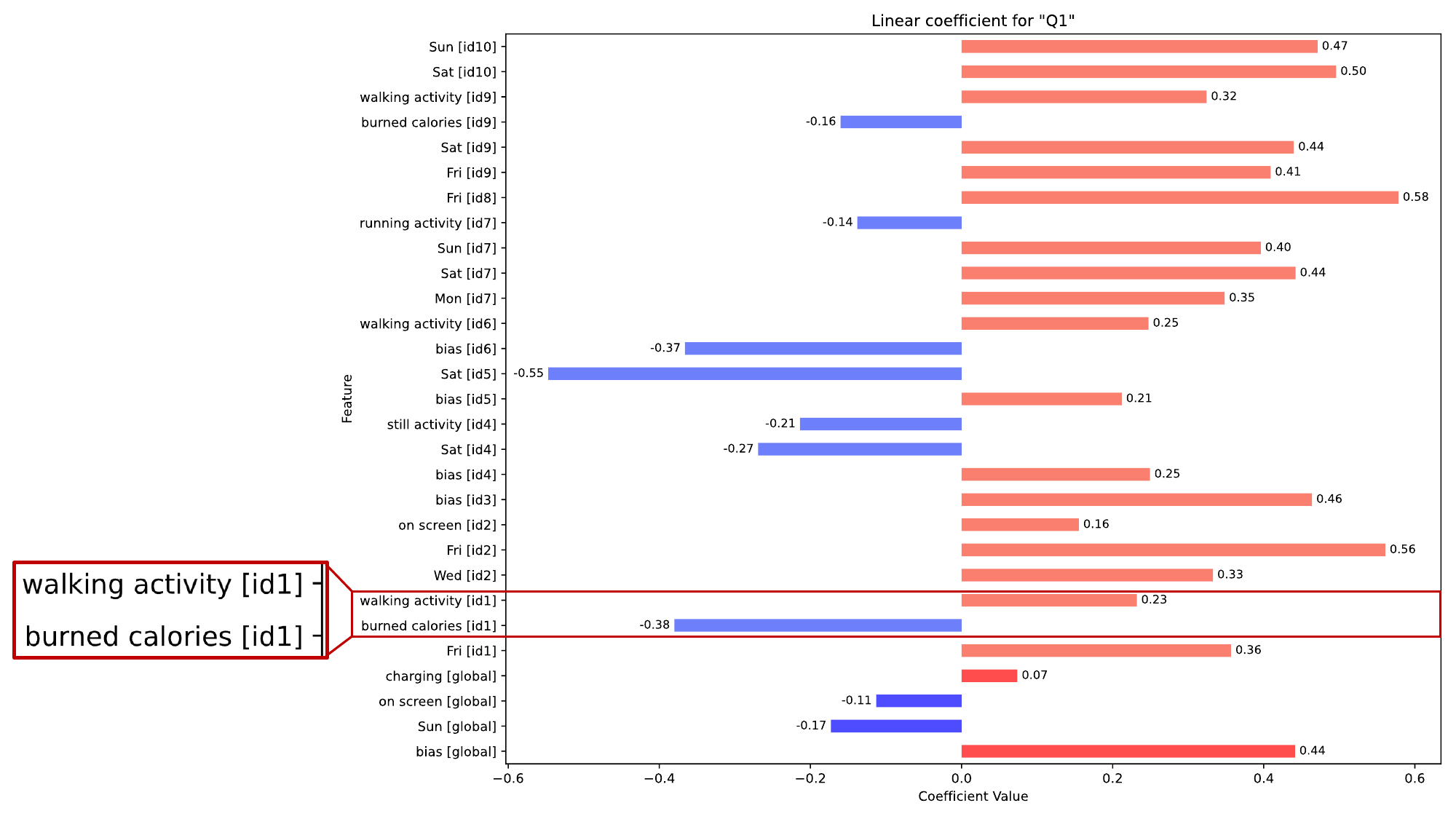}}
    \\
  \subfloat[Q2\label{fig:q2_coef}]{%
    \includegraphics[height=2.3cm, keepaspectratio]{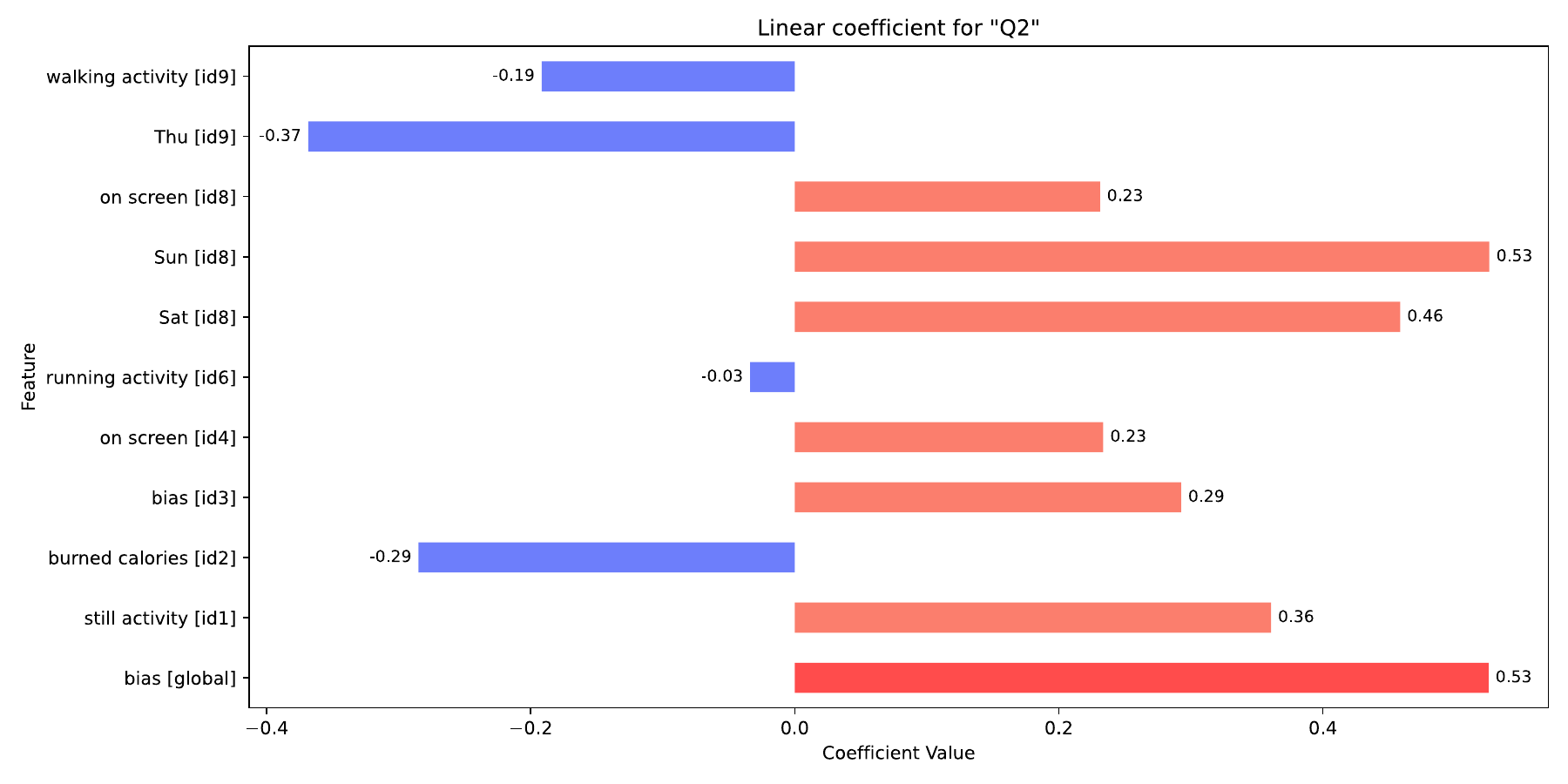}}
  \\
  \subfloat[Q3\label{fig:q3_coef}]{%
    \includegraphics[height=3.55cm, keepaspectratio]{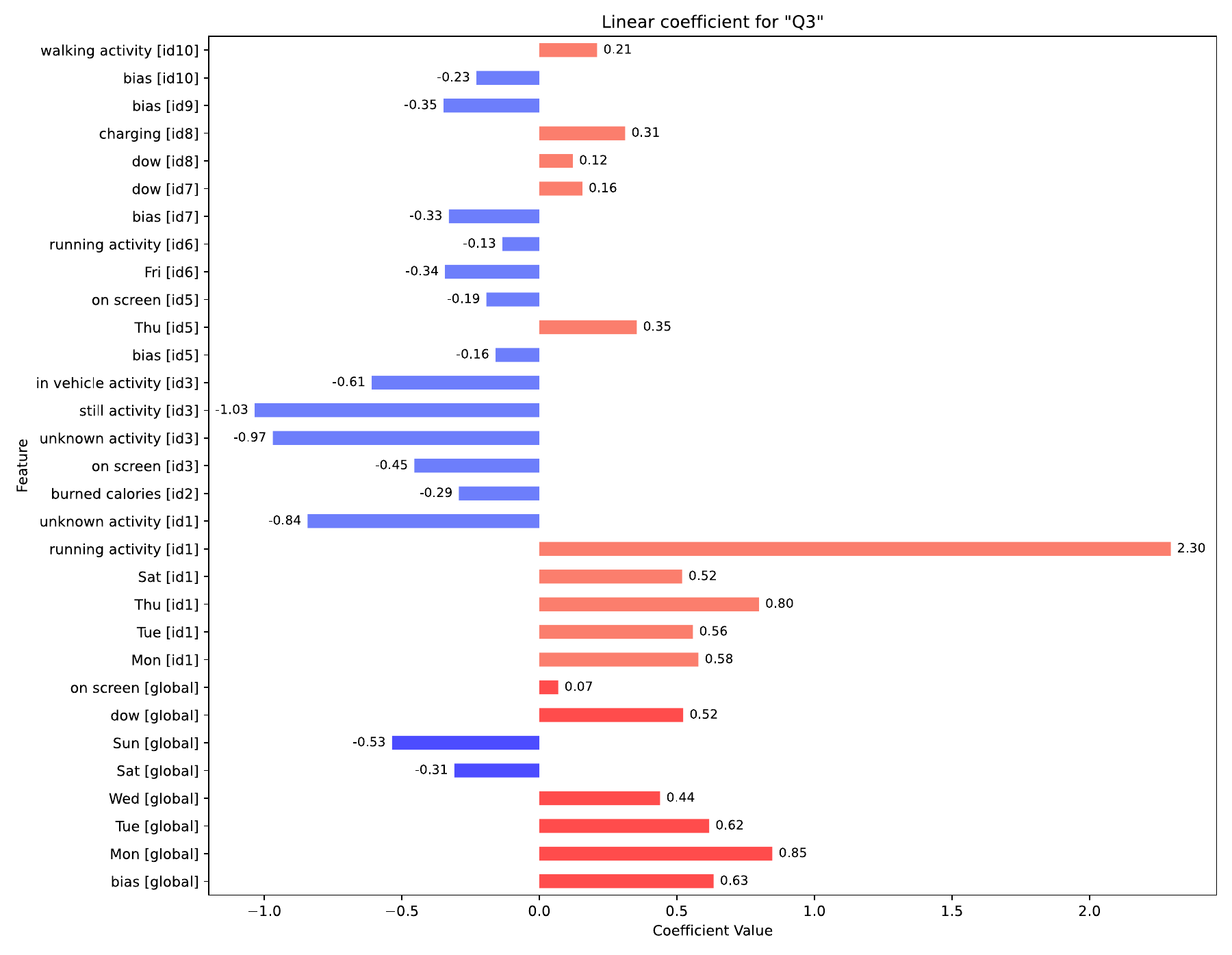}}
  \\
  \subfloat[S1\label{fig:s1_coef}]{%
    \includegraphics[height=2.3cm, keepaspectratio]{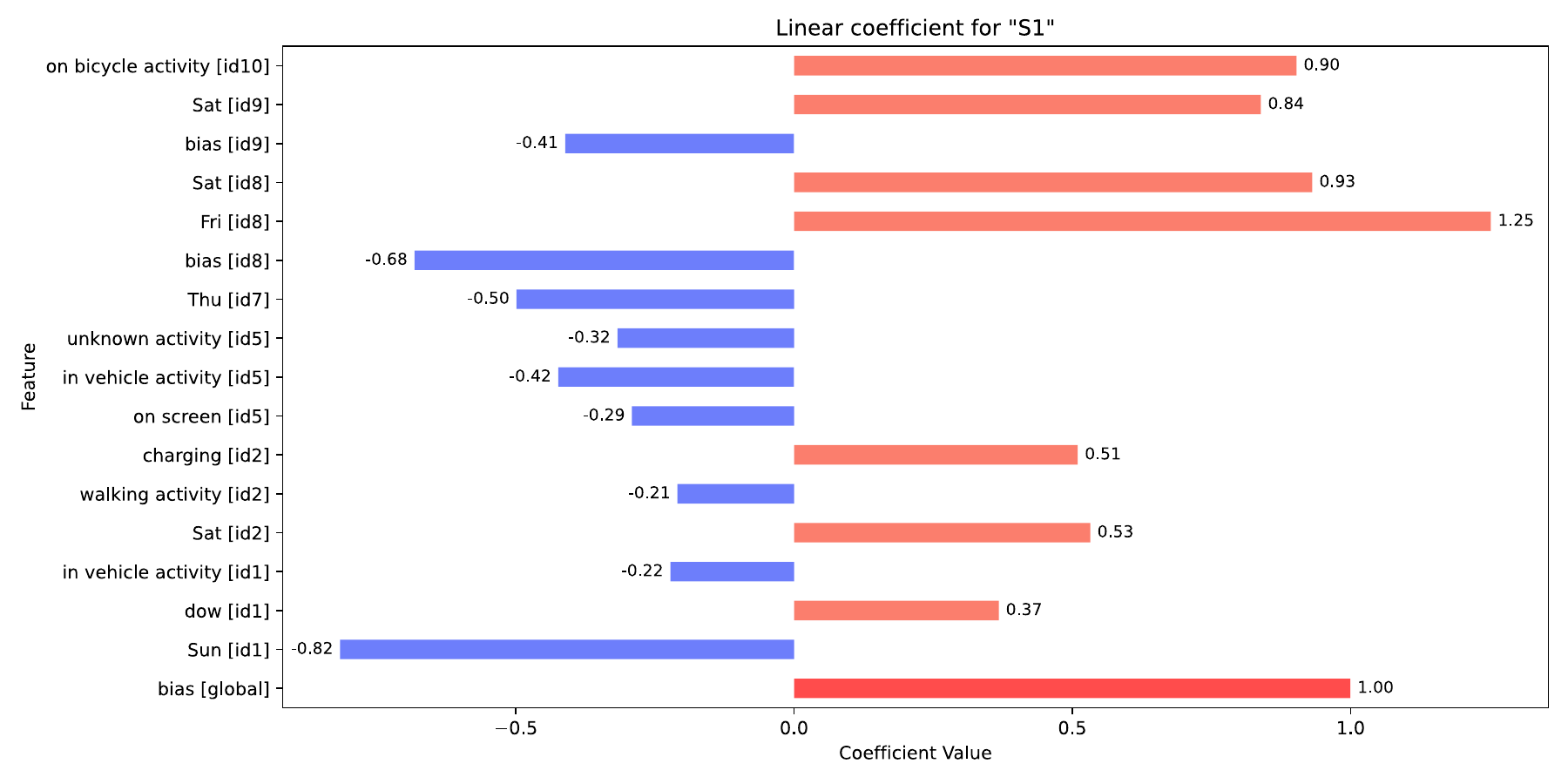}}
  \\
  \subfloat[S3\label{fig:s3_coef}]{%
    \includegraphics[height=2.7cm, keepaspectratio]{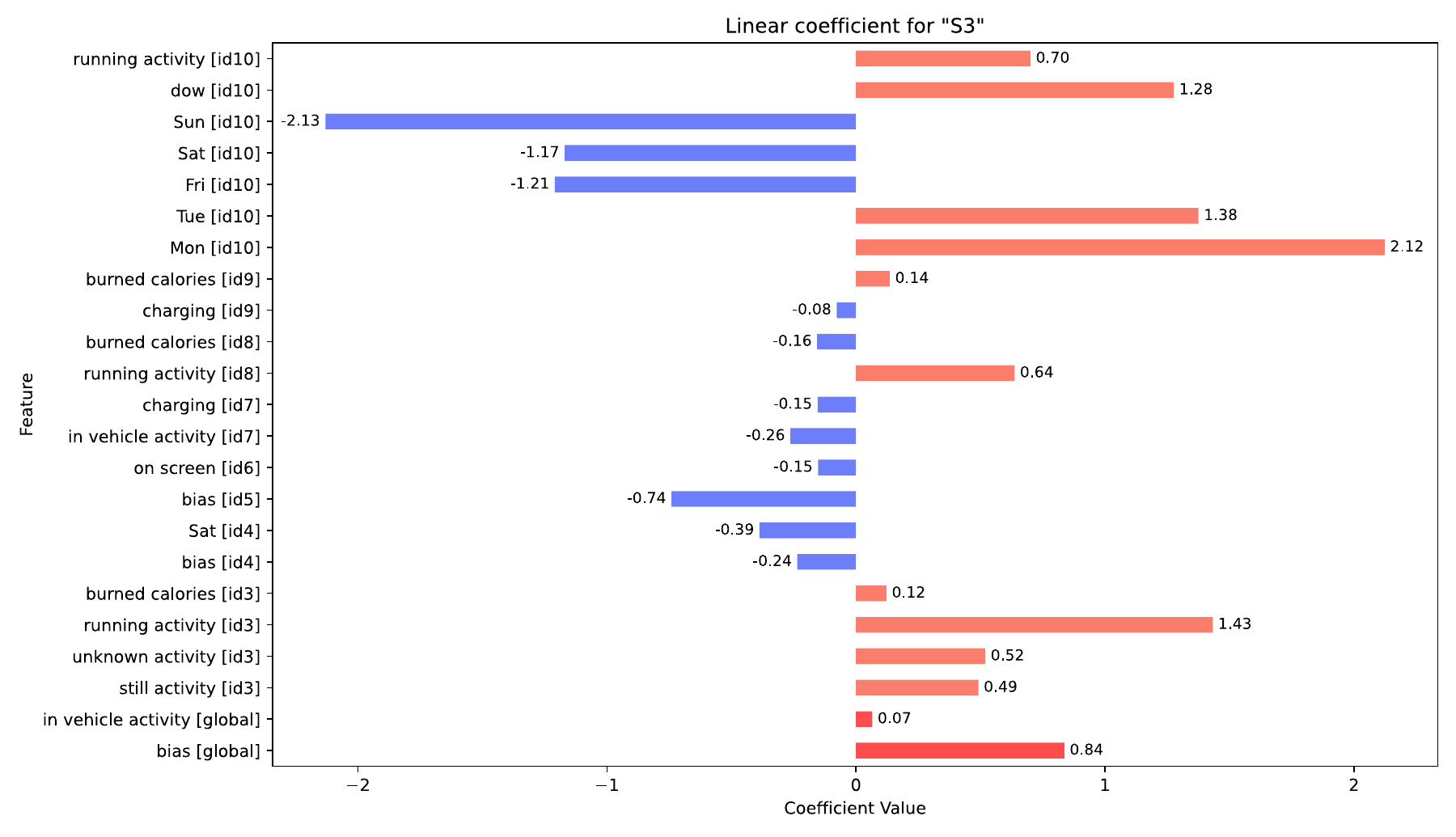}}
  \caption{Linear coefficient profiles for each target.}
  \label{fig:coef_grid}
\end{figure}

Lastly, interpretability remained central to SASL’s practical appeal. 
Figures~\ref{fig:q1_coef}–\ref{fig:s3_coef} illustrate how each linear model naturally decomposed into a global backbone that captures common patterns---such as the negative effects of excessive screen use and irregular charging on sleep quality (Q1)---along with subject-specific adjustments reflecting personal variations. 
For example, user \texttt{id01} showed a significantly negative weight for daily calorie expenditure, suggesting that excessive energy use could reduce this individual's perceived sleep quality the following morning. 
Conversely, a positive weight for walking activity implies that moderate physical activity could benefit \texttt{id01}'s overall perceived sleep quality. 
Thus, the SASL approach consistently enabled clear, actionable interpretations that linked model predictions directly to observable daily behaviors.

\section{Conclusion}\label{sec:con}
This study introduced the Subject-Adaptive Sparse Linear (SASL) framework, an interpretable linear modeling approach tailored explicitly for personalized health prediction tasks using multimodal lifelog data. 
By systematically separating global and subject-specific effects, employing rigorous backward feature elimination, and utilizing a regression-then-thresholding strategy optimized for ordinal macro-F1 evaluation, SASL effectively captured inter-individual variability while maintaining sparsity and transparency. 
Additionally, selectively integrating a compact LightGBM model through confidence-based gating further enhanced prediction accuracy without significantly compromising interpretability.

Nevertheless, avenues for future research remain. 
First, since the competition provided discretized labels, analyses using the original continuous or more finely graded ordinal labels could yield richer insights. 
Second, the current feature set predominantly consisted of daily summary statistics; future models incorporating detailed temporal and sequential dynamics could reveal deeper behavioral insights. 

Despite these considerations, SASL exhibited notable strengths: achieving interpretability comparable to classical linear models, predictive accuracy on par with sophisticated black-box methods, and significantly fewer parameters. 
Achieving a top-1\% placement on the public leaderboard further underscores the potential of interpretable, sparse modeling frameworks in personalized health prediction contexts.

\section*{Acknowledgment}

This work was supported in part by the National Research Foundation of Korea (NRF) funded by the Korean Government (Ministry of Science and ICT (MSIT)) under Grant RS-2024-00361377, and in part by the Institute for Information and Communications Technology Planning \& Evaluation (IITP) funded by the Korean Government (MSIT) under Grant RS-2025-02214591.

\medskip
{
\small
\bibliographystyle{IEEEtran}
\bibliography{ictc_ref}
}

\end{document}